\documentclass[10pt,twocolumn,letterpaper]{article}

\usepackage{iccv}
\usepackage{times}
\usepackage{epsfig}
\usepackage{graphicx}
\usepackage{amsmath}
\usepackage{amssymb}

\usepackage[breaklinks=true,bookmarks=false]{hyperref}

\iccvfinalcopy 

\def\iccvPaperID{****} 

\begin{document}

\title{Encouraging LSTMs to Anticipate Actions Very Early}

\author{Mohammad Sadegh Aliakbarian$^{1,3}$, Fatemeh Sadat Saleh$^{1,3}$, Mathieu Salzmann$^{2}$, Basura Fernando$^{1}$, \\Lars Petersson$^{1,3}$, Lars Andersson$^{3}$\\
$^{1}$Australian National University, $^{2}$CVLab, EPFL, Switzerland, $^{3}$Smart Vision Systems, CSIRO\\
{\tt\small firstname.lastname@data61.csiro.au, mathieu.salzmann@epfl.ch, basura.fernando@anu.edu.au}
}

\maketitle

\begin{abstract}
 In contrast to the widely studied problem of recognizing an action given a complete sequence, action anticipation aims to identify the action from only partially available videos. As such, it is therefore key to the success of computer vision applications requiring to react as early as possible, such as autonomous navigation. In this paper, we propose a new action anticipation method that achieves high prediction accuracy even in the presence of a very small percentage of a video sequence. To this end, we develop a multi-stage LSTM architecture that leverages context-aware and action-aware features, and introduce a novel loss function that encourages the model to predict the correct class as early as possible. Our experiments on standard benchmark datasets evidence the benefits of our approach; We outperform the state-of-the-art action anticipation methods for early prediction by a relative increase in accuracy of 22.0\% on JHMDB-21, 14.0\% on UT-Interaction and 49.9\% on UCF-101.
\end{abstract}


\section{Introduction}

Understanding actions from videos is key to the success of many real-world applications, such as autonomous navigation and sports analysis. While great progress has been made to recognize actions from complete sequences~\cite{SpaceTimeInterest,IDT,LRCN,DiscriminativeRankPooling,MultiStream,TSN,MultiStreamBiLSTM,RankPooling,DynamicNetwork} in the past decade, action anticipation~\cite{ryoo2011human,ryoo2009spatio,ryoo2010overview,yu2012predicting,soomro2016online,soomro2016predicting,ma2016learning}, which aims to predict the observed action as early as possible, has become a popular research problem only recently. 
Anticipation is crucial in scenarios where one needs to react before the action is finalized, such as to avoid hitting pedestrians with an autonomous car, or to forecast dangerous situations in surveillance scenarios.



\begin{figure}
\centering
\includegraphics[width=0.4\textwidth]{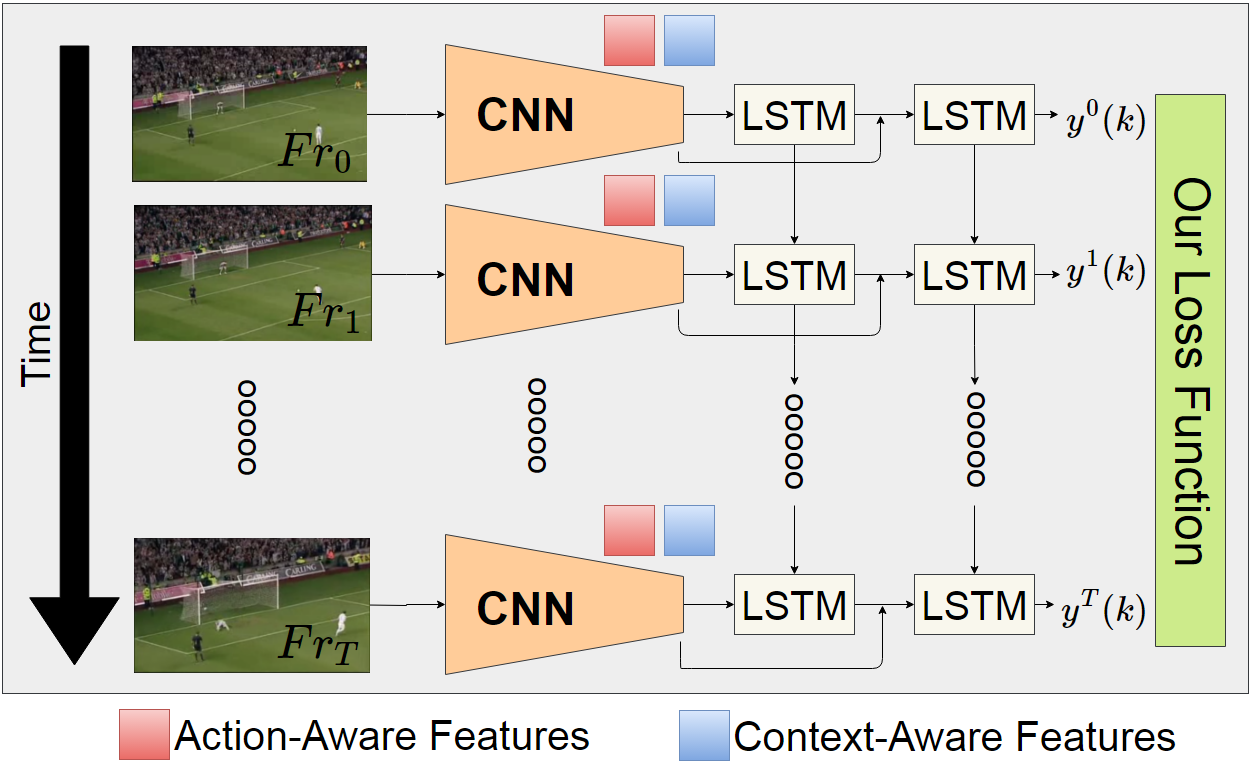}
\caption{
{\bf Overview of our approach.} Given a small portion of sequential data, our approach is able to predict the action category with very high performance. For instance, in UCF-101, our approach anticipates actions with more than 80\% accuracy given only the first 1\% of the video. To achieve this, we design a a model that leverages action- and context-aware features together with a new loss function that encourages the model to make correct predictions as early as possible.
}
\label{FIG:LOGO}
\end{figure}

The key difference between recognition and anticipation lies in the fact that the methods tackling the latter should predict the correct class as early as possible, given only a few frames from the beginning of the video sequence. To address this, several approaches have introduced new training losses encouraging the score~\cite{soomro2016online} or the rank~\cite{ma2016learning} of the correct action to increase with time, or penalizing increasingly strongly the classification mistakes~\cite{brain4Cars}. In practice, however, the effectiveness of these losses remains limited for very early prediction, such as from 1\% of the sequence.

In this paper, we introduce a novel loss that encourages making correct predictions very early. Specifically, our loss models the intuition that some actions, such as running and high jump, are highly ambiguous after seeing only the first few frames, and false positives should therefore not be penalized too strongly in the early stages. By contrast, we would like to predict a high probability for the correct class as early as possible, and thus penalize false negatives from the beginning of the sequence. Our experiments demonstrate that, for a given model, our new loss yields significantly higher accuracy than existing ones on the task of early prediction.

In particular, in this paper, we also contribute a novel multi-stage Long Short Term Memory (LSTM) architecture for action anticipation. This model effectively extracts and jointly exploits context- and action-aware features (see Fig.~\ref{FIG:LOGO}). This is in contrast to existing methods that typically extract either global representations for the entire image~\cite{DeepCAMP,TSN,ActionTransformation,LRCN} or video sequence~\cite{3DCNN,LargeScaleCNN}, thus not focusing on the action itself, or localize the feature extraction process to the action itself via dense trajectories~\cite{TrajectoryPooled,IDT,DiscriminativeRankPooling}, optical flow~\cite{CNN2Stream,TSN,VLAD3} or actionness~\cite{actionness,ActionnessRanking,ActionTubelets,FastActionProposal,OnlineSEEDS,SpatioTemporalProposal}, thus failing to exploit contextual information. To the best of our knowledge, only two-stream networks~\cite{TwoStreamNIPS,CNN2Stream,cheron2015p,SpatioTemporalLSTM} have attempted to jointly leverage both information types by making use of RGB frames in conjunction with optical flow to localize the action. Exploiting optical flow, however, does not allow these methods to explicitly leverage appearance in the localization process.
Furthermore, computing optical flow is typically expensive, thus significantly increasing the runtime of these methods. By not relying on optical flow, our method is significantly more efficient: On a single GPU, our model analyzes a short video (e.g., 50 frames) 14 times faster than~\cite{TwoStreamNIPS} and~\cite{CNN2Stream}.

Our model is depicted in Fig.~\ref{FIG:LSTM}. In a first stage, it focuses on the global, context-aware information by extracting features from the entire RGB image. The second stage then combines these context-aware features with action-aware ones obtained by exploiting class-specific activations, typically corresponding to regions where the action occurs. In short, our model first extracts the contextual information, and then merges it with the localized one. 

As evidenced by our experiments, our approach significantly outperforms the state-of-the-art action anticipation methods on all the standard benchmark datasets that we evaluated on, including UCF-101~\cite{UCF101}, UT-Interaction~\cite{ryoo2010overview}, and JHMDB21~\cite{JHMDB}. In the supplementary material, we further show that our combination of context- and action-aware features is also beneficial for the more traditional task of action recognition. Moreover, we evaluate the effect of optical flow features for both  action recognition and anticipation.

\section{Related Work}


The focus of this paper is twofold: Action anticipation, with a new loss that encourages correct prediction as early as possible, and action modeling, with a model that combines context- and action-aware information using multi-stage LSTMs. Below, we discuss the most relevant approaches for these two aspects.


\subsection{Action Anticipation}


The idea of action anticipation was introduced by~\cite{ryoo2009spatio}, which models causal relationships to predict human activities.
This was followed by several attempts to model the dynamics of the observed actions, such as by introducing integral and dynamic bag-of-words~\cite{ryoo2011human}, using spatial-temporal implicit shape models~\cite{yu2012predicting}, extracting human body movements via skeleton information~\cite{zhao2013online}, and accounting for the complete and partial history of observed features~\cite{kong2014discriminative}.


More recently,~\cite{soomro2016predicting} proposed to make use of binary SVMs to classify video snippets into sub-action categories and obtain the final class label in an online manner using dynamic programming. To overcome the need to train one classifier per sub-action,~\cite{soomro2016online} extended this approach to using a structural SVM. Importantly, this work further introduced a new objective function to encourage the score of the correct action to increase as time progresses.

While the above-mentioned work made use of handcrafted features, recent advances have naturally led to the development of deep learning approaches to action anticipation. In this context,~\cite{ma2016learning} proposed to combine a Convolutional Neural Network (CNN) with an LSTM to model both spatial and temporal information. The authors further introduced new ranking losses whose goal is to enforce either the score of the correct class or the margin between the score of the correct class and that of the best score to be non-decreasing over time. Similarly, in~\cite{brain4Cars}, a new loss that penalizes classification mistakes increasingly strongly over time was introduced in an LSTM-based framework that used multiple modalities. While the two above-mentioned methods indeed aim at improving classification accuracy over time, they do not explicitly encourage making correct predictions as early as possible. By contrast, while accounting for ambiguities in early stages, our new loss still aims to prevent false negatives from the beginning of the sequence.

Instead of predicting the future class label, in~\cite{vondrick2016anticipating}, the authors proposed to predict the future visual representation. However, the main motivation for this was to work with unlabeled videos, and the learned representation is therefore not always related to the action itself. 

\begin{figure*}[t!]
\centering
\includegraphics[width=0.9\textwidth]{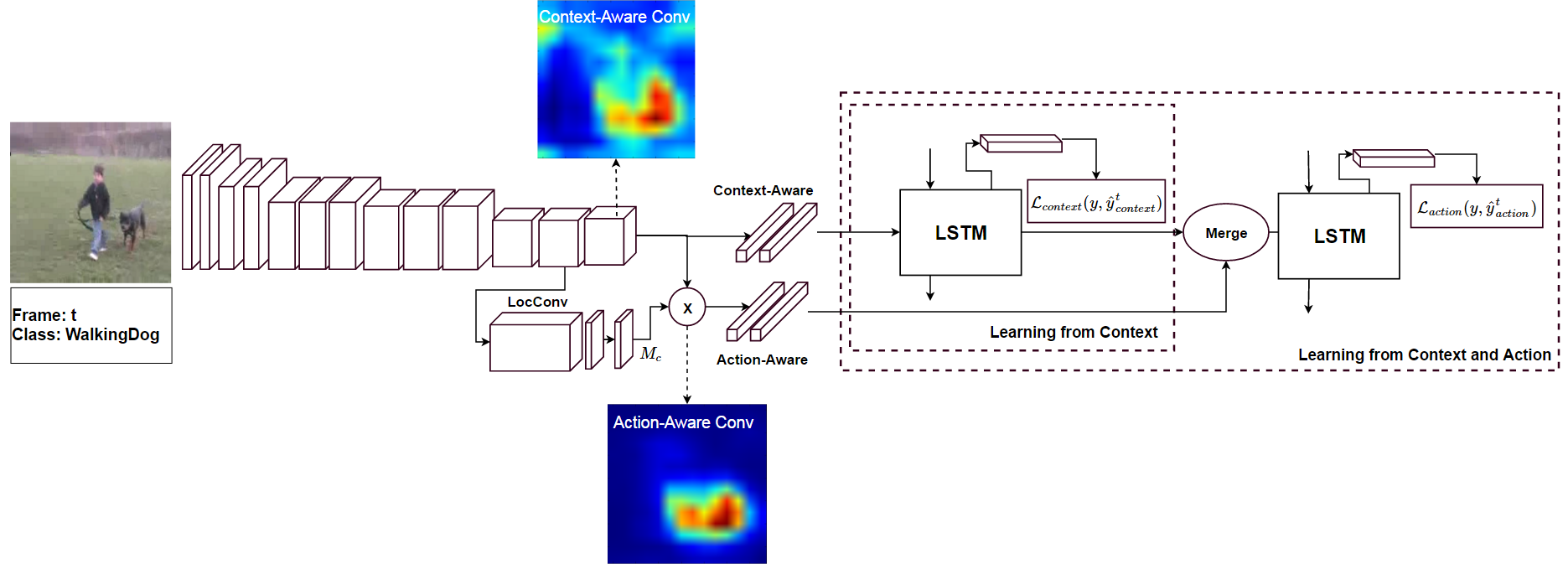}
\caption{{\bf Overview of our approach.} We propose to extract context-aware features, encoding global information about the scene, and combine them with action-aware ones, which focus on the action itself. To this end, we introduce a multi-stage LSTM architecture that leverages the two types of features to predict the action or forecast it. Note that, for the sake of visualization, the color maps were obtained from 3D tensors (\(512\times W\times H\)) via an average pooling operation over the 512 channels.
}
\label{FIG:LSTM}
\end{figure*}

\subsection{Action Modeling}
Most recent action approaches extract global representations for the entire image~\cite{DeepCAMP,ActionTransformation,LRCN} or video sequence~\cite{3DCNN,LargeScaleCNN}. As such, these methods do not truly focus on the actions of interest, but rather compute a \emph{context-aware} representation. Unfortunately, context does not always bring reliable information about the action. For example, one can play guitar in a bedroom, a concert hall or a yard. 
To overcome this, some methods localize the feature extraction process by exploiting dense trajectories~\cite{TrajectoryPooled,IDT,DiscriminativeRankPooling} or optical flow~\cite{VLAD3}. 
Inspired by objectness, the notion of actionness~\cite{actionness,ActionnessRanking,ActionTubelets,FastActionProposal,OnlineSEEDS,SpatioTemporalProposal} has recently also been proposed to localize the regions where a generic action occurs. The resulting methods can then be thought of as extracting \emph{action-aware} representations. In other words, these methods go to the other extreme and completely discard the notion of context which can be useful for some actions, such as playing football on a grass field.

There is nevertheless a third class of methods that aim to leverage these two types of information~\cite{TwoStreamNIPS,CNN2Stream,cheron2015p,SpatioTemporalLSTM,TSN}. By combining RGB frames and optical flow in two-stream architectures, these methods truly exploit context and motion, from which the action can be localized by learning to distinguish relevant motion. This localization, however, does not directly exploit appearance. Here, inspired by the success of these methods, we develop a novel multi-stage network that also leverages context- and action-aware information. However, we introduce a new action-aware representation that exploits the RGB data to localize the action. As such, our approach effectively leverages appearance for action-aware modeling, and, by avoiding the expensive optical flow computation, is much more efficient than the above-mentioned two-stream models.
In particular, our model is about 14 times faster than the state-of-the-art two-stream network of~\cite{CNN2Stream} and has less parameters. The reduction in number of parameters is due to the fact that~\cite{CNN2Stream} and~\cite{TwoStreamNIPS} rely on two VGG-like networks (one for each stream) with a few additional layers (including 3D convolutions for [8]). By contrast, our model has, in essence, a single VGG-like architecture, with some additional LSTM layers, which only have few parameters. Moreover, our work constitutes the first attempt at explicitly leveraging context- and action-aware information for action anticipation. Finally, we introduce a novel multi-stage LSTM fusion strategy to integrate action and context aware features.

\section{Our Approach}
Our goal is to predict the class of an action as early as possible, that is, after having seen only a very small portion of the video sequence. To this end, we first introduce a new loss function that encourages making correct predictions very early. We then develop a multi-stage LSTM model that makes use of this loss function while leveraging both context- and action-aware information.

\subsection{A New Loss for Action Anticipation}
\label{sec:loss}
As argued above, a loss for action anticipation should encourage having a high score for the correct class as early as possible. However, it should also account for the fact that, early in the sequence, there can often be ambiguities between several actions, such as running and high jump. Below, we introduce a new anticipation loss that follows these two intuitions.

Specifically, let $y^t(k)$ encode the true activity label at time $t$, i.e., $y^t(k) = 1$ if the sample belongs to class $k$ and 0 otherwise, and $\hat{y}^t(k)$ denote the corresponding label predicted by a given model. We define our new loss as
\begin{align}
\mathcal{L}(y, \hat{y}) = -\frac{1}{N}\sum^N_{k=1}\sum^T_{t=1}\Bigg[y^t(k) \log(\hat{y}^t(k)) + \nonumber \\ \frac{t(1-y^t(k))}{T}\log(1-\hat{y}^t(k))\Bigg]\;,
\label{eq:loss}
\end{align}
where $N$ is the number of action classes and $T$ the length (number of frames) of the input sequence.

This loss function consists of two terms. The first one penalizes false negatives with the same strength at any point in time. By contrast, the second one focuses on false positives, and its strength increases linearly over time, to reach the same weight as that on false negatives.
Therefore, the relative weight of the first term compared to the second one is larger at the beginning of the sequence. Altogether, this encourages predicting a high score for the correct class as early as possible, i.e., preventing false negatives, while accounting for the potential ambiguities at the beginning of the sequence, which give rise to false positives. As we see more frames, however, the ambiguities are removed, and these false positives are encouraged to disappear.


Our new loss matches the desirable properties of an action anticipation loss. In the next section, we introduce a novel multi-stage architecture that makes use of this loss.

\subsection{Multi-stage LSTM Architecture}
To tackle action anticipation, we develop the novel multi-stage recurrent architecture based on LSTMs depicted by Fig.~\ref{FIG:LSTM}. This architecture consists of a stage-wise combination of context- and action-aware information. Below, we first discuss our approach to extracting these two types of information, and then present our complete multi-stage recurrent network.


\begin{figure}
\centering
\includegraphics[width=0.45\textwidth]{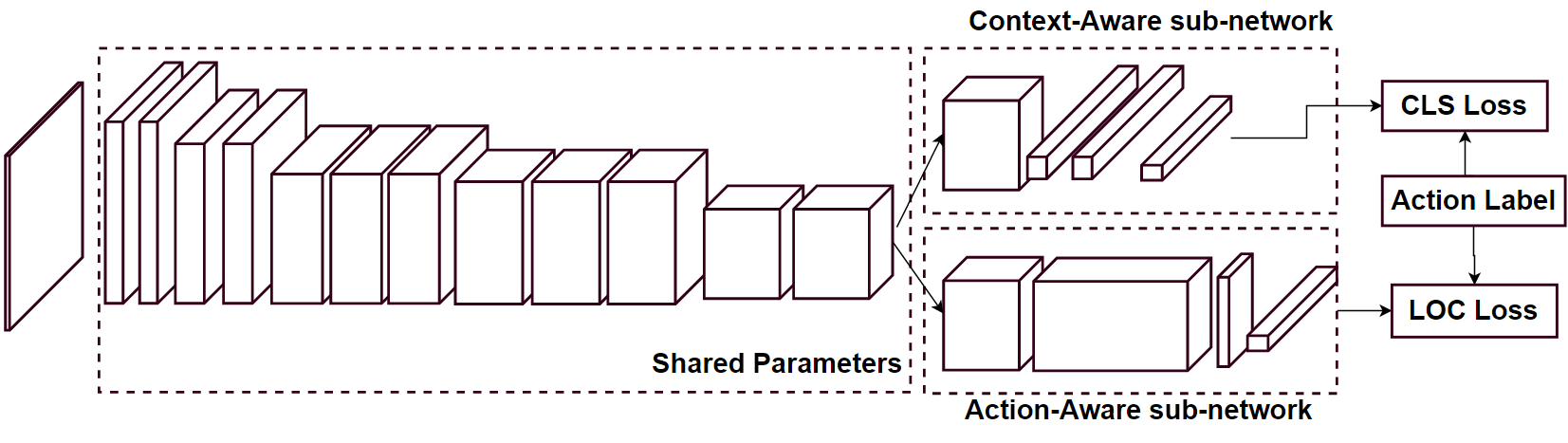}
\caption{{\bf Our feature extraction network.} Our CNN model for feature extraction is based on the VGG-16 structure with some modifications. Up to conv5-2, the network is the same as VGG-16. The output of this layer is connected to two sub-models. The first one extracts context-aware features by providing a global image representation. The second one relies on another network to extract action-aware features.}
\label{FIG:LOC_TRAIN}
\end{figure}

\subsubsection{Context- and Action-aware Modeling}
To model the context- and action-aware information, we introduce the two-stream architecture shown in Fig.~\ref{FIG:LOC_TRAIN}. The first part of this network is shared by both streams and, up to conv5-2, corresponds to the VGG-16 network~\cite{VGG}, pre-trained on ImageNet for object recognition. The output of this layer is connected to two sub-models: One for context-aware features and the other for action-aware ones. We then train these two sub-models for the same task of action recognition from a single image using a cross-entropy loss function defined on the output of each stream. In practice, we found that training the entire model in an end-to-end manner did not yield a significant improvement over training only the two sub-models. In our experiments, we therefore opted for this latter strategy, which is less expensive computationally and memory-wise. Below, we first discuss the context-aware sub-network and then turn to the action-aware one.

\vspace{-0.4cm}
\paragraph{Context-Aware Feature Extraction.}
This sub-model is similar to VGG-16 from conv5-3 up to the last fully-connected layer, with the number of units in the last fully-connected layer changed from 1000 (original 1000-way ImageNet classification model) to the number of activities $N$. 

In essence, this sub-model focuses on extracting a deep representation of the whole scene for each activity and thus incorporates context. We then take the output of its fc7 layer as our context-aware features.

\paragraph{Action-Aware Feature Extraction.}
As mentioned before, the context of an action does not always correlate with the action itself. Our second sub-model therefore aims at extracting features that focus on the action itself. To this end, we draw inspiration from the object classification work of~\cite{zhou2015learning}. At the core of this work lies the idea of Class Activation Maps (CAMs). In our context, a CAM indicates the regions in the input image that contribute most to predicting each class label. In other words, it provides information about the location of an action. Importantly, this is achieved without requiring any additional annotations.

More specifically, CAMs are extracted from the activations in the last convolutional layer in the following manner. Let \(f_l(x,y)\) represent the activation of unit \(l\) in the last convolutional layer at spatial location \((x,y)\). A score $S_k$ for each class $k$ can be obtained by performing global average pooling~\cite{NetinNetGAP} to obtain, for each unit $l$, a feature \(F^l = \sum_{x,y}{f_l(x,y)}\), followed by a linear layer with weights $\{w_l^k\}$. That is, \(S_k = \sum_k{w_l^k F_l}\). A CAM for class $k$ at location \((x,y)\) can then be computed as
\begin{equation}
M_k(x,y) = \sum_l{w_l^k f_l(x,y)}\;,
\end{equation}
which indicates the importance of the activations at location \((x,y)\) in the final score for class $k$.

\begin{figure}
\centering
\includegraphics[width=0.45\textwidth]{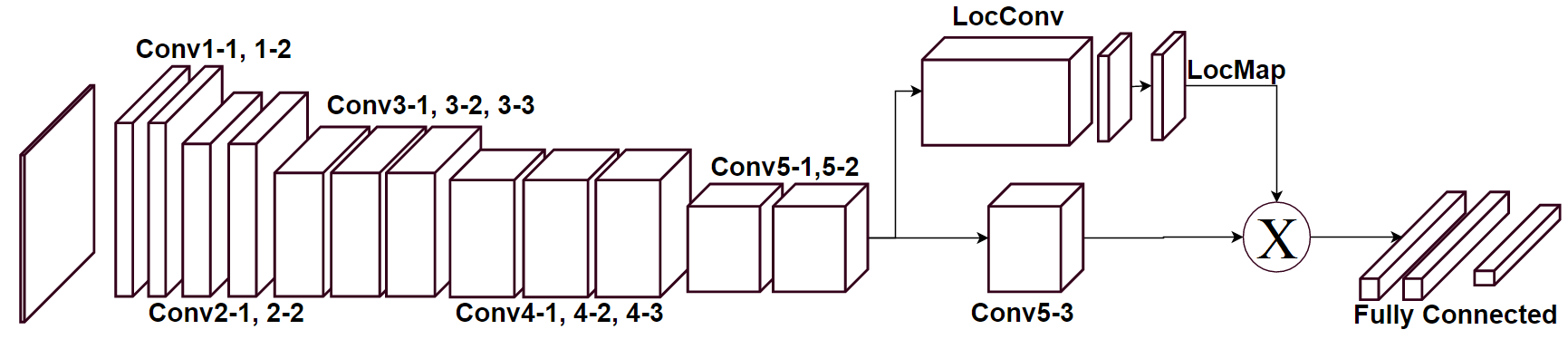}
\caption{{\bf Action-aware feature extraction.} Given the fine-tuned feature extraction network, we introduce a new layer that alters the output of conv5-3. This lets us filter out the conv5-3 features that are irrelevant, to focus on the action itself. Our action-aware features are taken as the output of the last fully-connected layer.}
\label{FIG:LOC_TEST}
\end{figure}

Here, we propose to make use of the CAMs to extract action-aware features. To this end, we use the CAMs in conjunction with the output of the conv5-3 layer of the model. The intuition behind this is that conv5-3 extracts high-level features that provide a very rich representation of the image~\cite{UnderstandingCNN} and typically correspond to the most discriminative parts of the object~\cite{DeepEdge,BuiltinFGBG}, or, in our case, the action. Therefore, we incorporate a new layer to our sub-model, whose output can be expressed as
\begin{equation}
A_k(x,y) = {\rm conv_{5-3}}(x,y) \times {\rm ReLU}(M_k(x,y))\;,
\end{equation}
where ${\rm ReLU}(M_k(x,y)) = {\rm max}(0, M_k(x,y))$. As shown in Fig.~\ref{FIG:LOC_TEST}, this new layer is then followed by fully-connected ones, and we take our action-aware features as the output of the corresponding fc7 layer.

\subsubsection{Sequence Learning for Action Anticipation}
\label{sec:seq_learn}
To effectively combine the information contained in the context-aware and action-aware features described above, we design the novel multi-stage LSTM model depicted by Fig.~\ref{FIG:LSTM}. This model first focuses on the context-aware features, which encode global information about the entire image. It then combines the output of this first stage with our action-aware features to provide a refined class prediction.


To train this model for action anticipation, we make use of our new loss introduced in Section~\ref{sec:loss}. Therefore, ultimately, our network models long-range temporal information and yields increasingly accurate predictions as it processes more frames. 

Specifically, we write the overall loss of our model as
\begin{equation}
\mathcal{L}_o = \frac{1}{V} \sum_{i=1}^V \left(\mathcal{L}_{c,i} + \mathcal{L}_{a,i}\right)\;,
\label{eq.lstm.overall}
\end{equation}
where $V$ is the total number of training sequences. This loss function combines losses corresponding to the context-aware stage and to the action-aware stage, respectively.
Below, we discuss these two stages in more detail.

\paragraph{Learning Context.}
The first stage of our model takes as input our context-aware features, and passes them through a layer of LSTM cells followed by a fully-connected layer that, via a softmax operation, outputs a probability for each action class. Let $\hat{y}_{c,i}$ be the vector of probabilities for all classes and all time steps predicted by the first stage for sample $i$. We then define the loss for a single sample as 
\begin{equation}
\mathcal{L}_{c,i} = \mathcal{L}(y_{i},\hat{y}_{c,i})\;,
\end{equation}
where $\mathcal{L}(\cdot)$ is our new loss defined in Eq.~\ref{eq:loss}, and $y_{i}$ is the ground-truth class label for sample $i$.


\paragraph{Learning Context and Action.}
The second stage of our model aims at combining context-aware and action-aware information. Its structure is the same as that of the first stage, i.e., a layer of LSTM cells followed by a fully-connected layer to output class probabilities via a softmax operation. However, its input merges the output of the first stage with our action-aware features. This is achieved by concatenating the hidden activations of the LSTM layer with our action-aware features. We then make use of the same loss function as before, but defined on the final prediction. For sample $i$, this can be expressed as
\begin{equation}
\mathcal{L}_{a,i} = \mathcal{L}(y_i, \hat{y}_{{a,i}})\;,
\end{equation}
where $\hat{y}_{a}$ is the vector of probabilities for all classes predicted by the second stage.


\paragraph{Inference.}
At inference, the input RGB frames are forward-propagated through our model. We therefore obtain a probability vector for each class at each frame. While one could simply take the probabilities in the current frame $t$ to obtain the class label at time $t$, via $argmax$, we propose to increase robustness by leveraging the predictions of all the frames up to time $t$. To this end, we make use of an average pooling of these predictions over time.

\section{Experiments}
In this section, we first compare our method with state-of-the-art techniques on the task of action anticipation, and then analyze various aspects of our model, such as the influence of the loss function and of the different feature types. In the supplementary material, we provide additional experiments to analyze the effectiveness of different LSTM architectures, and the influence of the number of hidden units and of our temporal average pooling strategy. We also report the performance of our method on the task of action recognition from complete videos with and without optical flow, and action anticipation with optical flow.

\subsection{Datasets}
For our experiments, we made use of the standard UCF-101~\cite{UCF101}, UT-Interaction~\cite{ryoo2010overview}, and JHMDB-21~\cite{JHMDB} benchmarks, which we briefly describe below.

The UCF-101 dataset consists of 13,320 videos (each contains a single action) of 101 action classes including a broad set of activities such as sports, playing musical instruments and human-object interaction, with an average length of 7.2 seconds. UCF-101 is one of the most challenging datasets due to its large diversity in terms of actions and to the presence of large variations in camera motion, cluttered background and illumination conditions. There are three standard training/test splits for this dataset. In our comparisons to the state-of-the-art for both action anticipation and recognition, we report the average accuracy over the three splits. For the detailed analysis of our model, however, we rely on the first split only.

The JHMDB-21 dataset is another challenging dataset of realistic videos from various sources, such as movies and web videos, containing 928 videos and 21 action classes. Similarly to UCF-101, in our comparison to the state-of-the-art, we report the average accuracy over the three standard splits of data. Similar to UCF-101 dataset, each video contains one action starting from the beginning of the video.

The UT-Interaction dataset contains videos of continuous executions of 6 human-human interaction classes: shake-hands, point, hug, push, kick and punch. 
The dataset contains 20 video sequences whose length is about 1 minute each. Each video contains at least one execution of each interaction type, providing us with 8 executions of human activities per video on average. Following the recommended experimental setup, we used 10-fold leave-one-out cross validation for each of the standard two sets of 10 videos. That is, within each set, we leave one sequence for testing and use the remaining 9 for training. 
Following standard practice, we also made use of the annotations provided with the dataset to split each video into sequences containing individual actions.


\subsection{Implementation Details}
\paragraph{CNN and LSTM Configuration.}
The parameters of the CNN were optimized using stochastic gradient descent with a fixed learning rate of 0.001, a momentum of 0.9, a weight decay of 0.0005, and mini-batches of size 32. To train our LSTMs, we similarly used stochastic gradient descent with a fixed learning rate of 0.001, a momentum of 0.9, and mini-batch size of 32. For all LSTMs, we used 2048 hidden units. To implement our method, we used Python and Keras~\cite{keras}. We will make our code publicly available. 

\paragraph{Training Procedure.}
To fine-tune the network on each dataset, we augment the data, so as to reduce the effect of over-fitting. The input images were randomly flipped horizontally and rotated by a random amount in the range -8 to 8 degrees. We then extracted crops according to the following procedure:
\textbf{(1)} Compute the maximum cropping rectangle with a given aspect ratio ($320/240$) that fits inside the input image.
\textbf{(2)} Scale the width and height of the cropping rectangle by a factor randomly selected in the range $0.8$-$1$.
\textbf{(3)} Select a random location for the cropping rectangle within the original input image and extract the corresponding subimage.
\textbf{(4)} Scale the subimage to $224 \times 224$.

After these geometric transformations, we further applied RGB channel shifting~\cite{wu2015deep}, followed by randomly adjusting image brightness, contrast and saturation with a factor $\alpha=0.3$. The operations are: for brightness, $\alpha \times Image$, for contrast, $Image \times \alpha + (1.0 - \alpha)\times mean(grey(Image))$, and for saturation, $Image\times \alpha + (1.0 - \alpha)\times grey(Image)$. 


\subsection{Comparison to the State-of-the-Art}
We compare our approach to the state-of-the-art action anticipation results reported on each of the three datasets discussed above.
We further complement these state-of-the-art results with additional baselines that make use of our context-aware features with the loss of either~\cite{ma2016learning} or~\cite{brain4Cars}. Note that a detailed comparison of different losses within our model is provided in Section~\ref{sec:res_loss}.

Following standard practice, we report the so-called \emph{earliest} and \emph{latest} prediction accuracies. Note, however, that there is no real agreement on the proportion of frames that the \emph{earliest} setting corresponds to. For each dataset, we make use of the proportion that has been employed by the baselines (i.e., either 20\% or 50\%). Note also that our approach relies on at most $T$ frames (with $T=50$ in practice). Therefore, in the \emph{latest} setting, where the baselines rely on the complete sequences, we only exploit the first $T$ frames. We believe that the fact that our method significantly outperforms the state-of-the-art in this setting despite using less information further evidences the effectiveness of our approach.

\paragraph{JHMDB-21.}
The results for the JHMDB-21 dataset are provided in Table~\ref{tab:jhmdb}. In this case, following the baselines, earliest prediction corresponds to observing the first 20\% of the sequence. Note that we clearly outperform all the baselines by a significant margin in both the earliest and latest settings. Remarkably, we also outperform the methods that rely on additional information as input, such as optical flow~\cite{soomro2016online,soomro2016predicting,ma2016learning} and Fisher vector features based on Improved Dense Trajectories~\cite{soomro2016online}. This clearly demonstrates the benefits of our approach for anticipation.

\begin{table}
\centering
\small
\caption{Comparison with state-of-the-art baselines on the task of action anticipation on the JHMDB-21 dataset. Note that our approach outperforms all baselines significantly in both settings.
}
\label{tab:jhmdb}
\begin{tabular}{l c c c}
\hline
Method & Earliest & Latest\\
\hline
DP-SVM~\cite{soomro2016online} & 5\% & 46\% \\
S-SVM~\cite{soomro2016online} & 5\% & 43\% \\
Where/What~\cite{soomro2016predicting} & 10\% & 43\%\\
Ranking Loss~\cite{ma2016learning} & 29\% & 43\% \\
Context-Aware+Loss of~\cite{brain4Cars} & 28 \% & 43\% \\
Context-Aware+Loss of~\cite{ma2016learning} & 33\% & 39\% \\
\hline
Ours & \textbf{55\%} & \textbf{58\%} \\
\hline
\end{tabular}
\end{table}

\begin{table}
\centering
\small
\caption{Comparison with state-of-the-art baselines on the task of action anticipation on the UT-Interaction dataset.}
\label{tab:ut}
\begin{tabular}{l c c c}
\hline
Method & Earliest & Latest \\
\hline
D-BoW~\cite{ryoo2011human} & 70.0\% & 85.0\% \\
I-BoW~\cite{ryoo2011human} & 65.0\% & 81.7\% \\
CuboidSVM~\cite{ryoo2010overview} & 31.7\% & 85.0\% \\
BP-SVM~\cite{laviers2009improving} & 65.0\% & 83.3\% \\
CuboidBayes~\cite{ryoo2011human} & 25.0\% & 71.7\% \\
DP-SVM~\cite{soomro2016online} & 13.0\% & 14.6\%\\
S-SVM~\cite{soomro2016online} & 11.0\% & 13.4\% \\
Context-Aware+Loss of~\cite{brain4Cars} & 45.0 \% & 65.0\% \\
Context-Aware+Loss of~\cite{ma2016learning} & 48.0\% & 60.0\% \\
\hline
Ours & \textbf{84.0\%} & \textbf{90.0\%}\\
\hline
\end{tabular}
\end{table}

\begin{table}
\centering
\small
\caption{Action Anticipation on the UCF-101 dataset. 
}
\label{tab:ucf}
\begin{tabular}{l c c c}
\hline
Method & Earliest & Latest \\
\hline
Context-Aware+Loss of~\cite{brain4Cars} & 30.6 \% & 71.1\% \\
Context-Aware+Loss of~\cite{ma2016learning} & 22.6\% & 73.1\% \\
\hline
Ours & \textbf{80.5\%} & \textbf{83.4\%} \\
\hline
\end{tabular}
\end{table}

\paragraph{UT-Interaction.}
We provide the results for the UT-Interaction dataset in Table~\ref{tab:ut}. Here, following standard practice, 50\% of  the sequence was observed for earliest prediction, and the entire sequence for latest prediction. Recall that our approach uses at most $T=50$ frames for prediction in both settings, while the average length of a complete sequence is around 120 frames. Therefore, as evidenced by the results, our approach yields significantly higher accuracy despite using considerably less data as input.





\paragraph{UCF-101.}
We finally compare our approach with our two baselines on the UCF-101 dataset. While this is not a standard benchmark for action anticipation, this experiment is motivated by the fact that this dataset is relatively large, has many classes, with similarity across different classes, and contains variations in video capture conditions. Altogether, this makes it a challenging dataset to anticipate actions, especially when only a small amount of data is available. The results on this dataset are provided in Table~\ref{tab:ucf}. Here, the earliest setting corresponds to using the first 2 frames of the sequences, which corresponds to around 1\% of the data. Again, we clearly outperform the two baselines consisting of exploiting context-aware features with the loss of either~\cite{ma2016learning} or~\cite{brain4Cars}. We believe that this further evidences the benefits of our approach, which leverages both context- and action-aware features with our new anticipation loss. A detailed evaluation of the influence of the different feature types and losses is provided in the next section.


\subsection{Analysis}
In this section, we provide a more detailed analysis of the influence of our loss function and of the different feature types on anticipation accuracy. Finally, we also provide a visualization of our different feature types, to illustrate their respective contributions.


\begin{figure}
\centering
\includegraphics[width=0.44\textwidth]{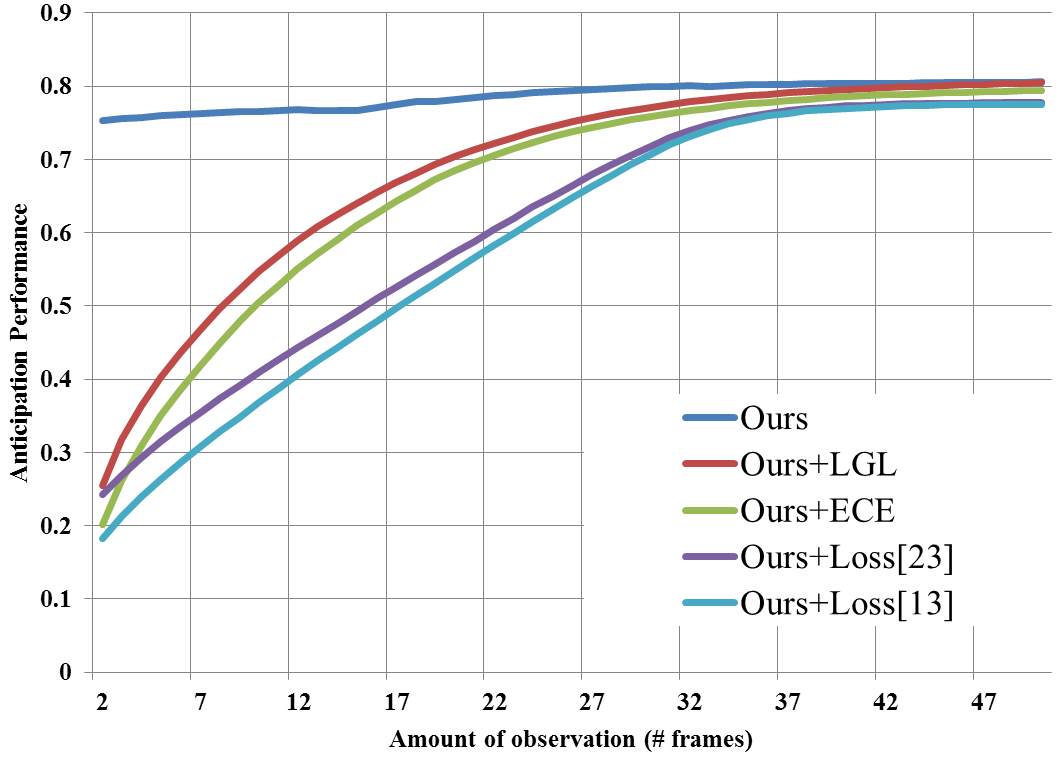}
\caption{{\bf Comparison of different losses for action anticipation on UCF-101.} We evaluate the accuracy of our model trained with different losses as a function of the number of frames observed.
This plot clearly shows the superiority of our loss function.}
\label{fig:anticipation}
\end{figure}

\begin{figure}
\centering
\includegraphics[width=0.44\textwidth]{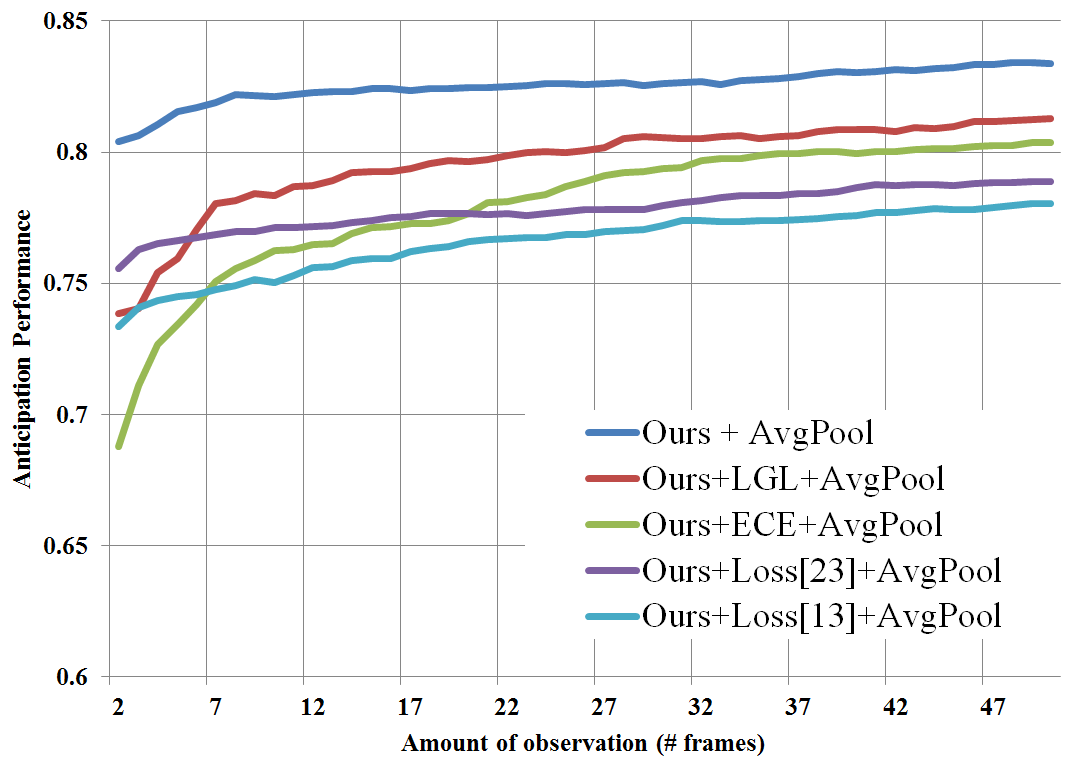}
\caption{{\bf Influence of our average pooling strategy.} Our simple yet effective average pooling leverages the predictions of all the frames up to time $t$. As shown on the UCF-101 dataset, this increases anticipation performance, especially at very early stages. 
}
\label{fig:anticipation_pool}
\end{figure}

\subsubsection{Influence of the Loss Function}
\label{sec:res_loss}
Throughout the paper, we have argued that our novel loss, introduced in Section~\ref{sec:loss}, is better-suited to action anticipation than existing ones. To evaluate this, we trained several versions of our model with different losses. In particular, as already done in the comparison to the state-of-the-art above, we replaced our loss with the ranking loss of~\cite{ma2016learning} (ranking loss on detection score) 
and the loss of~\cite{brain4Cars}, but this time within our complete multi-stage model, with both context- and action-aware features.

Furthermore, we made use of the standard cross-entropy ({\bf CE}) loss, which only accounts for one activity label for each sequence (at time $T$). This loss can be expressed as
\begin{align}
\mathcal{L}_{CE} = \sum^{N}_{k=1}[y^T(k)\log(\hat{y}^{T}(k))\ \nonumber \\+ (1-y^T(k))\log(1-\hat{y}^{T}(k))]\;.
\end{align}

We then also modified the loss of~\cite{brain4Cars}, which consists of an exponentially weighted softmax, with an exponentially weighted cross-entropy loss ({\bf ECE}),
written as
\begin{align}
\mathcal{L}_{ECE} = \sum^T_{t=1}-e^{-(T-t)}\sum^{N}_{k=1}[y^t(k)\log(\hat{y}^{t}(k)) \nonumber \\+ (1-y^t(k))\log(1-\hat{y}^{t}(k))]\;.
\end{align}

The main drawback of this loss comes from the fact that it does not strongly encourage the model to make correct predictions as early as possible. To address this issue, we also introduce a linearly growing loss  ({\bf LGL}), defined as
\begin{align}
\mathcal{L}_{LGL} = \sum^T_{t=1}{-\frac{t}{T}}\sum^{N}_{k=1}[y^t(k)\log(y^{t}(k)) \nonumber \\ + (1-y^t(k))\log(1-\hat{y}^{t}(k))].
\end{align}
While our new loss, introduced in Section~\ref{sec:loss}, also makes use of a linearly-increasing term, it corresponds to the false positives in our case, as opposed to the false negatives in the LGL. Since some actions are ambiguous in the first few frames, we find it more intuitive not to penalize false positives too strongly at the beginning of the sequence. This intuition is supported by our results below, which show that our loss yields better results than the LGL.

In Fig.~\ref{fig:anticipation}, we report the accuracy of the corresponding models as a function of the number of observed frames on the UCF-101 dataset. Note that our new loss yields much higher accuracies than the other ones, particularly when only a few frames of the sequence are observed; With only 2 frames observed, our loss yields an accuracy similar to the other losses with 30--40 frames. With 30fps, this essentially means that we can predict the action 1 second earlier than other methods. The importance of this result is exemplified by research showing that a large proportion of vehicle accidents are due to mistakes/misinterpretations of the scene in the immediate time leading up to the crash~\cite{brain4Cars,crash}. 


Moreover, in Fig.~\ref{fig:anticipation_pool}, we report the performance of the corresponding models as a function of the number of observed frames when using our average pooling strategy. Note that this strategy can generally be applied to any action anticipation method and, as shown by comparing Figs.~\ref{fig:anticipation} and~\ref{fig:anticipation_pool}, increases accuracy, especially at very early stages, which clearly demonstrates its effectiveness. Note that using it in conjunction with our loss still yields the best results by a significant margin.



\subsubsection{Influence of the Features}
We then evaluate the importance of the different feature types, context-aware and action-aware, on anticipation accuracy. To this end, we compare models trained using each feature type individually with our model that uses them jointly. For the models using a single feature type, we made use of a single LSTM to model temporal information. By contrast, our approach relies on a multi-stage LSTM, which we denote by \emph{MS-LSTM}. Note that all models were trained using our new anticipation loss. The results of this experiment on the UCF-101 dataset are provided in Table~\ref{tab:features}. These results clearly evidence the importance of using both feature types, which consistently outperforms individual ones.

\begin{table}
\renewcommand{\arraystretch}{1.2}
\small
\centering
\caption{Importance of the different feature types.}
\label{tab:features}
\begin{tabular}{l l c c c}
\hline
Feature & Model  & Earliest \scriptsize{K=1}& Latest \scriptsize{K=50}\\
\hline
Context-Aware 	& LSTM& 62.80\%& 72.71\%\\
Action-Aware  	& LSTM & 69.33\%& 77.86\% \\
Context+Action 	& MS-LSTM	& 80.5\%	& 83.37\%	\\
\hline
\end{tabular}
\end{table}

Since we extract action-aware features, and not motion-aware ones, our approach will not be affected by irrelevant motion. The CNN that extracts these features learns to focus on the discriminative parts of the images, thus discarding the irrelevant information. To confirm this, we conducted an experiment on some classes of UCF-101 that contain irrelevant motions/multiple actors, such as Baseball pitch, Basketball, Cricket Shot and Ice dancing. The results of our action-aware and context-aware frameworks for these classes are: 66.1\% vs. 58.2\% for Baseball pitch, 83\% vs. 76.4\% for Basketball, 65.1\% vs. 58\% for Cricket Shot, and 92.3\% vs. 91.7\% for Ice dancing. This shows that our action-aware features can effectively discard irrelevant motion/actors to focus on the relevant one(s).

\subsubsection{Visualization}
\begin{figure}
\centering
\small
\begin{tabular}{c c c c}
\scriptsize{Apply Lipstick} & \scriptsize{Band Marching} & \scriptsize{Archery} & \scriptsize{Blowing Candles} \\
\includegraphics[width=0.081\textwidth]{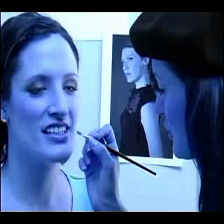} & 
\includegraphics[width=0.081\textwidth]{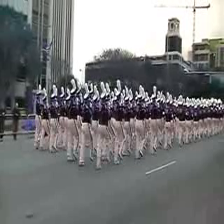} & 
\includegraphics[width=0.081\textwidth]{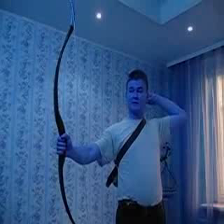} & 
\includegraphics[width=0.081\textwidth]{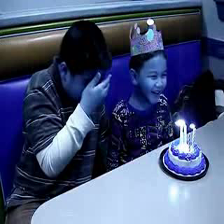} \\
\multicolumn{4}{c}{\scriptsize{RGB Frame}}\\
\includegraphics[width=0.081\textwidth]{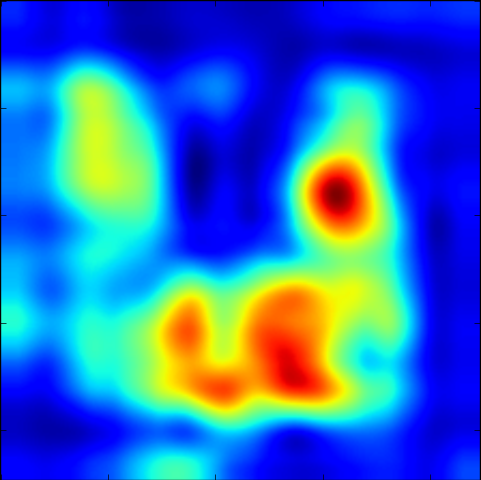} & 
\includegraphics[width=0.081\textwidth]{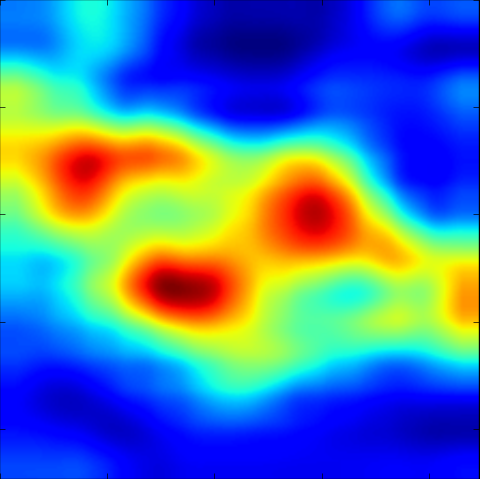} & 
\includegraphics[width=0.081\textwidth]{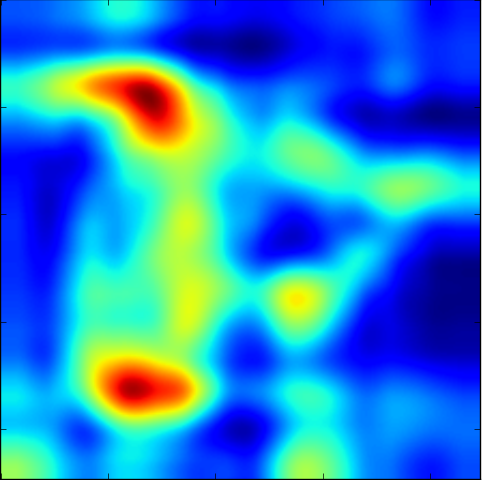} & 
\includegraphics[width=0.081\textwidth]{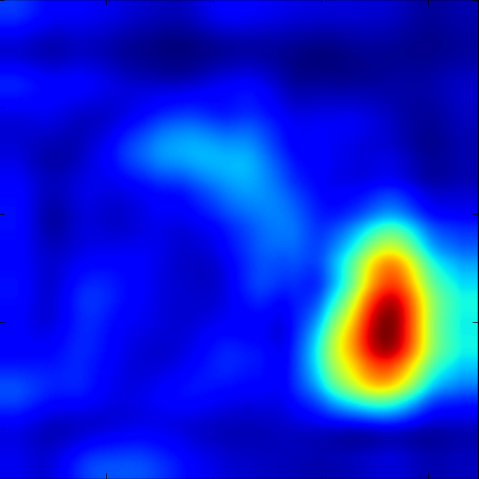} \\
\multicolumn{4}{c}{\scriptsize{Context-Aware Conv5}}\\
\includegraphics[width=0.081\textwidth]{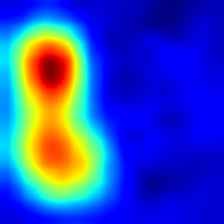} & 
\includegraphics[width=0.081\textwidth]{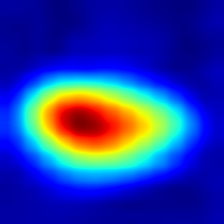} & 
\includegraphics[width=0.081\textwidth]{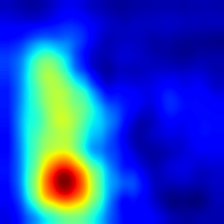} & 
\includegraphics[width=0.081\textwidth]{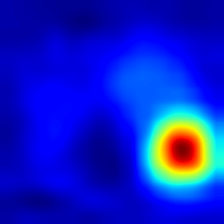} \\
\multicolumn{4}{c}{\scriptsize{Action-Aware Conv5}}\\
\end{tabular}
\caption{Visualization of action-aware and context-aware features on UCF-101. These samples are representative of the data.}
\label{fig:visulaization}
\end{figure}

Finally, we provide a better intuition of the kind of information each of our feature types encode (see Fig.~\ref{fig:visulaization}). 
This visualization was computed by average pooling over the 512 channels of Conv5-3 (of both the context-aware and action-aware sub-networks). As can be observed in the figure, our context-aware features have high activations on regions corresponding to any relevant object in the scene (context). By contrast, in our action-aware features, high activations correctly correspond to the focus of the action. Therefore, they can reasonably localize the parts of the frame that most strongly participate in the action happening in the video and reduce the noise coming from context.

\section{Conclusion}
In this paper, we have introduced a novel loss function to address very early action anticipation. Our loss encourages the model to make correct predictions as early as possible in the input sequence, thus making it particularly well-suited to action anticipation. Furthermore, we have introduced a new multi-stage LSTM model that effectively combines context-aware and action-aware features. Our experiments have evidenced the benefits of our new loss function over existing ones. Furthermore, they have shown the importance of exploiting both context- and action-aware information. Altogether, our approach significantly outperforms the state-of-the-art in action anticipation on all the datasets we applied it to. In the future, we intend to study new ways to incorporate additional sources of information, such as dense trajectories and human skeletons in our framework.

{\small
\bibliographystyle{ieee}
\bibliography{egbib}
}

\newpage
\pagebreak

\section*{Supplementary Material}
In this supplementary material, we analyze different aspects of our approach via several additional experiments. While the main paper discusses action anticipation, here, we focus on evaluating our approach on the task of action recognition. Therefore, we first provide a comparison to the state-of-the-art action recognition methods on three standard benchmarks, and evaluate the effect of exploiting additional optical flow features for both action recognition and anticipation. We then analyze the effect of our different feature types in several loss functions, the influence of the number of hidden units and of our average pooling in LSTMs, and, finally, the effect of our multi-stage LSTM architecture.

\section*{Comparison to State-of-the-Art Action Recognition Methods}
We first compare the results of our approach to state-of-the-art methods on UCF-101, JHMDB-21 and UT-Interaction in terms of average accuracy over the standard training and testing partitions. In Table~\ref{tab:soa_recognition_ucf}, we provide the results on the UCF-101 dataset. Here, for the comparison to be fair, we only report the results of the baselines that do not use any other information than the RGB image and the activity label (we refer the readers to the baselines' papers and the survey~\cite{survey} for more detail). In other words, while it has been shown that additional, handcrafted features, such as dense trajectories and optical flow, can help improve accuracy~\cite{IDT,TrajectoryPooled,VLAD3,TwoStreamNIPS,DynamicNetwork}, our goal here is to truly evaluate the benefits of our method, not of these features. Note, however, that, as discussed in the next section of this supplementary material, our approach can still benefit from such features. As can be seen from the table, our approach outperforms all these RGB-based baselines. In Tables~\ref{tab:soa_recognition_jhmdb} and~\ref{tab:soa_recognition_uti}, we provide the results for JHMDB-21 and UT-Interaction. Again, we outperform all the baselines, even though, in this case, some of them rely on additional information such as optical flow~\cite{FindingActionTubes,actionness,soomro2016online,soomro2016predicting,ma2016learning} or IDT Fisher vector features~\cite{soomro2016online}.
We believe that these experiments show the effectiveness of our approach at tackling the action recognition problem.

\begin{table}
\centering
\caption{Comparison with state-of-the-art methods on UCF-101 (average accuracy over all training/testing splits). For the comparison to be fair, we focus on the baselines that, as us, only use the RGB frames as input.}
\label{tab:soa_recognition_ucf}
\begin{tabular}{l c}
\hline
Method & Accuracy\\
\hline
Dynamic Image Network~\cite{DynamicNetwork} & 70.0\% \\
Dynamic Image Network + Static RGB~\cite{DynamicNetwork} & 76.9\%\\
Rank Pooling~\cite{DiscriminativeRankPooling} & 72.2\%\\
DHR~\cite{DiscriminativeRankPooling} & 78.8\%\\
Zhang et al.~\cite{RealTimeAction} & 74.4\% \\
LSTM~\cite{LSTMAction} & 74.5\%\\
LRCN~\cite{LRCN} & 68.8\% \\
C3D~\cite{3DCNN} & 82.3\% \\
Spatial Stream Net~\cite{TwoStreamNIPS} & 73.0\% \\
Deep Network~\cite{LargeScaleCNN} & 65.4\% \\
ConvPool (Single frame)~\cite{BeyondAction} & 73.3\%  \\
ConvPool (30 frames)~\cite{BeyondAction} & 80.8\%\\
ConvPool (120 frames)~\cite{BeyondAction} & 82.6\% \\
\hline
Ours & {\bf 83.3\%}\\
\hline
Diff. to State-of-the-Art & +0.7\% \\
\hline
\end{tabular}
\end{table}

\begin{table}
\centering
\caption{Comparison with state-of-the-art methods on JHMDB-21 (average accuracy over all training/testing splits). Note that while the methods of~\cite{FindingActionTubes,actionness,soomro2016online,soomro2016predicting} use motion/optical flow information and~\cite{soomro2016online} uses IDT Fisher vector features, our method yields better performance.}
\label{tab:soa_recognition_jhmdb}
\begin{tabular}{l c}
\hline
Method & Accuracy \\
\hline
Where and What~\cite{soomro2016predicting} & 43.8\%\\
DP-SVM~\cite{soomro2016online} & 44.2\% \\
S-SVM~\cite{soomro2016online} & 47.3\%\\
Spatial-CNN~\cite{FindingActionTubes} & 37.9\% \\
Motion-CNN~\cite{FindingActionTubes} & 45.7\% \\
Full Method~\cite{FindingActionTubes} & 53.3\%\\
Actionness-Spatial~\cite{actionness} & 42.6\%\\
Actionness-Temporal~\cite{actionness} & 54.8\% \\
Actionness-Full Method~\cite{actionness} & 56.4\%\\
\hline
Ours& \bf{58.3\%} \\
\hline
Diff. to State-of-the-Art & +1.9\%\\
\hline
\end{tabular}
\end{table}

\begin{table}
\centering
\caption{Comparison with state-of-the-art methods on UT-Interaction (average accuracy over all training/testing splits). Note that while the methods of~\cite{soomro2016online} uses motion/optical flow information and IDT Fisher vector features, our method yields better performance.}
\label{tab:soa_recognition_uti}
\begin{tabular}{l c}
\hline
Method & Accuracy \\
\hline
D-BoW~\cite{ryoo2011human} & 85.0\%\\
I-BoW~\cite{ryoo2011human} & 81.7\%\\
Cuboid SVM~\cite{ryoo2010overview} & 85.0\%\\
BP-SVM~\cite{laviers2009improving} & 83.3\%\\
Cuboid/Bayesian~\cite{ryoo2011human} & 71.7\%\\
DP-SVM~\cite{soomro2016online} & 14.6\%\\
Yu et al.~\cite{yu2010real} & 83.3\%\\
Yuan et al.~\cite{yuan2010middle} & 78.2\%\\
Waltisberg et al.~\cite{waltisberg2010variations} & 88.0\%\\
\hline
Ours & \bf{90.0\%} \\
\hline
Diff. to State-of-the-Art & +2.0\%   \\
\hline
\end{tabular}
\end{table}

\section*{Exploiting Optical Flow}
\label{sec:flow}
Note that our approach can also be extended into a two-stream architecture to benefit from optical flow information, as state-of-the-art action recognition methods do. In particular, to extract optical flow features, we made use of the pre-trained temporal network of~\cite{TwoStreamNIPS}. We then computed the CNN features from a stack of 20 optical flow frames (10 frames in the $x$-direction and 10 frames in the $y$-direction), from $t-10$ to $t$ at each time $t$. As these features are potentially loosely related to the action (by focusing on motion), we merge them with the input to the second stage of our multi-stage LSTM. In Table~\ref{tab:opticalFlow}, we compare the results of our modified approach with state-of-the-art methods that also exploit optical flow. Note that our two-stream approach yields accuracy comparable to the state-of-the-art.


\begin{table}
\renewcommand{\arraystretch}{1.2}
\centering
\small
\caption{Comparison with the state-of-the-art approaches that use optical flow. For the comparison to be fair, we focus on the baselines that, as us, use RGB frames+optical flow as input.}
\label{tab:opticalFlow}
\begin{tabular}{l c }
\hline
Method & Accuracy \\
\hline
Spatio-temporal ConvNet~\cite{LargeScaleCNN}			& 65.4\% \\
LRCN + Optical Flow~\cite{LRCN}  										& 82.9\% \\
LSTM + Optical Flow~\cite{LSTMAction}									& 84.3\% \\
Two-Stream Fusion~\cite{CNN2Stream} & 92.5\% \\
CNN features + Optical Flow~\cite{TwoStreamNIPS} 		& 73.9\% \\
ConvPool (30 frames) + OpticalFlow~\cite{BeyondAction} 	& 87.6\%\\
ConvPool (120 frames) + OpticalFlow~\cite{BeyondAction} & 88.2\%\\
VLAD3 + Optical Flow~\cite{VLAD3} 						& 84.1\% \\
Two-Stream ConvNet~\cite{TwoStreamNIPS} 					& 88.0\% \\
Two-Stream Conv.Pooling~\cite{BeyondAction}				& 88.2\% \\
Two-Stream TSN~\cite{TSN} 											& 91.5\% \\

\hline
Ours + Optical Flow 	& 91.8\% \\
\hline
\end{tabular}
\end{table}

We also conducted an experiment to evaluate the effectiveness of incorporating optical flow in our framework for action anticipation. To handle the case where less than 10 frames are used, we padded the frame stack with gray images (with values 127.5). Our flow-based approach achieved 86.8\% for earliest and 91.8\% for latest prediction on UCF-101, thus showing that, if runtime is not a concern, optical flow can indeed help increase the accuracy of our approach.

We further compare our approach with the two-stream network~\cite{TwoStreamNIPS}, designed for action recognition, applied to the task of action anticipation. On UCF-101, this model achieved 83.2\% for earliest and 88.6\% for latest prediction, which our approach with optical flow clearly outperforms.

\section*{Effect of Different Feature Types}
Here, we evaluate the importance of the different feature types, context-aware and action-aware, on recognition accuracy. To this end, we compare models trained using each feature type individually with our model that uses them jointly. For all models, we made use of LSTMs with 2048 units. Recall that our approach relies on a multi-stage LSTM, which we denote by \emph{MS-LSTM}. The results of this experiment for different losses are reported in Table~\ref{tab:features}. These results clearly evidence the importance of using both feature types, which consistently outperforms using individual ones in all settings.

\begin{table}
\renewcommand{\arraystretch}{1.2}
\centering
\caption{Importance of the different feature types using different losses. Note that combining both types of features consistently outperforms using a single one. Note also that, for a given model, our new  loss yields higher accuracies than the other ones.}
\label{tab:features}
\begin{tabular}{l l c}
\hline
Feature & Sequence Learning & Accuracy \\
\hline
Context-Aware	& LSTM (CE)	& 72.38\%\\
Action-Aware 	& LSTM (CE) & 74.24\% \\
Context+Action 	& MS-LSTM (CE)	& 78.93\%\\
\hline
Context-Aware	& LSTM (ECE)& 72.41\% \\
Action-Aware 	& LSTM (ECE) & 77.20\% \\
Context+Action 	& MS-LSTM (ECE)	& 80.38\%\\
\hline
Context-Aware	& LSTM (LGL) & 72.58\%\\
Action-Aware 	& LSTM (LGL) & 77.63\% \\
Context+Action 	& MS-LSTM (LGL) & 81.27\%\\
\hline
Context-Aware	& LSTM (Ours)& 72.71\%\\
Action-Aware 	& LSTM (Ours) & 77.86\% \\
Context+Action 	& MS-LSTM (Ours)	& 83.37\%	\\
\hline
\end{tabular}
\end{table}

\section*{Robustness to the Number of Hidden Units}

Based on our experiments, we found that for large datasets such as UCF-101, the 512 hidden units that some baselines use (e.g.~\cite{LRCN,LSTMAction}) do not suffice to capture the complexity of the data. Therefore, to study the influence of the number of units in the LSTM, we evaluated different versions of our model with 1024 and 2048 hidden units (since 512 yields poor results and higher numbers, e.g., 4096, would require too much memory) and trained the model with 80\% training data and validated on the remaining 20\%. 
For a single LSTM, we found that using 2048 hidden units performs best. For our multi-stage LSTM, using 2048 hidden units also yields the best results. We also evaluated the importance of relying on average pooling in the LSTM. The results of these different versions of our MS-LSTM framework are provided in Table~\ref{tab:AvgPool}. This shows that, typically, more hidden units and average pooling can improve accuracy slightly.

\begin{table}
\renewcommand{\arraystretch}{1.2}
\centering
\small
\caption{Influence of the number of hidden LSTM units and of our average pooling strategy in our multi-stage LSTM model. These experiments were conducted on the first splits of UCF-101 and JHMDB-21.}
\label{tab:AvgPool}
\begin{tabular}{l  c c c c}
\hline
  & Average  & Hidden & & \\
Setup  & Pooling & Units & UCF-101 & JHMDB-21\\
\hline
Ours (CE)& wo/ & 1024 & 77.26\%	& 52.80\% \\
Ours (CE)& wo/ & 2048 & 78.09\%	& 53.43\% \\
Ours (CE)& w/ & 2048 &	78.93\% & 54.30\%\\
\\
Ours (ECE)& wo/ & 1024 & 79.10\%	& 55.33\% \\
Ours (ECE)& wo/ & 2048 & 79.41\%	& 56.12\% \\
Ours (ECE)& w/ & 2048 & 80.38\%	& 57.05\%\\
\\
Ours (LGL)& wo/ & 1024 & 79.76\%	& 55.70\% \\
Ours (LGL)& wo/ & 2048 & 80.10\%	& 56.83\% \\
Ours (LGL)& w/ & 2048 & 81.27\%	& 57.70\%\\
\\
Ours & wo/ & 1024 & 81.94\%	& 56.24\% \\
Ours & wo/ & 2048 & 82.16\%	& 57.92\%\\
Ours & w/ & 2048 & 83.37\%	& 58.41\%\\
\hline
\end{tabular}
\end{table}

\section*{Effect of the LSTM Architecture}
Finally, we study the effectiveness of our multi-stage LSTM architecture at merging our two feature types. To this end, we compare the results of our MS-LSTM with the following baselines: A single-stage LSTM that takes as input the concatenation of our context-aware and action-aware features (Concatenation); The use of two parallel LSTMs whose outputs are merged by concatenation and then fed to a fully-connected layer (Parallel). A multi-stage LSTM where the two different feature-types are processed in the reverse order (Swapped), that is, the model processes the action-aware features first and, in a second stage, combines them with the context-aware ones;
The results of this comparison are provided in Table~\ref{tab:LSTMArch}. Note that both multi-stage LSTMs outperform the single-stage one and the two parallel LSTMs, thus indicating the importance of treating the two types of features sequentially. Interestingly, processing context-aware features first, as we propose, yields higher accuracy than considering the action-aware ones at the beginning. This matches our intuition that context-aware features carry global information about the image and will thus yield noisy results, which can then be refined by exploiting the action-aware features.

\begin{table}
\renewcommand{\arraystretch}{1.2}
\centering
\caption{Comparison of our multi-stage LSTM model with diverse fusion strategies. We report the results of simple concatenation of the context-aware and action-aware features, their use in two parallel LSTMs with late fusion, and swapping their order in our multi-stage LSTM, i.e., action-aware first, followed by context-aware. Note that multi-stage architectures yield better results, with the best ones achieved by using context first, followed by action, as proposed in this paper.}
\label{tab:LSTMArch}
\begin{tabular}{l l c}
\hline
Feature & Sequence & \\
Order & Learning & Accuracy \\
\hline
Concatenation 	& LSTM  				& 77.16\% \\
Parallel 		& 2 Parallel LSTMs  	& 78.63\% \\
Swapped 		& MS-LSTM (Ours)  				& 78.80\% \\
Ours 			& MS-LSTM (Ours) 			& 83.37\% \\
\hline
\end{tabular}
\end{table}

Furthermore, we evaluate a CNN-only version of our approach, where we removed the LSTM, but kept our average pooling strategy to show the effect of our MS-LSTM architecture on top of the CNN. On UCF-101, this achieved 69.53\% for earliest and 73.80\% for latest prediction. This shows that, while this CNN-only framework yields reasonable predictions, our complete approach with our multistage LSTM benefits from explicitly being trained on multiple frames, thus achieving significantly higher accuracy (80.5\% and 83.4\%, respectively). While the LSTM could in principle learn to perform average pooling, we believe that the lack of data prevents this from happening. 

\paragraph{Acknowledgement.} Authors thank Oscar Friberg for his assistance in conducting additional experiments of the supplementary material.

\end{document}